\documentclass{svproc}
\usepackage{url}

\usepackage{amssymb}
\usepackage{graphicx}
\usepackage{makecell}
\begin{document}
\mainmatter              % start of a contribution
\title{Imbalanced Sample Generation and Evaluation for Power System Transient Stability Using CTGAN}
\titlerunning{Imbalanced Sample Generation and Evaluation}  % abbreviated title (for running head)
\author{Gengshi Han \and Shunyu Liu \and Kaixuan Chen \and Na Yu \and Zunlei Feng \and Mingli Song}
\authorrunning{Gengshi Han et al.} % abbreviated author list (for running head)
\tocauthor{Gengshi Han, Shunyu Liu, Kaixuan Chen, Na Yu, Zunlei Feng, and Mingli Song}
\institute{Zhejiang University, Hangzhou, 310007, Zhejiang Province, China\\
\email{\{hangengshi, liushunyu, chenkx, na\_yu, zunleifeng, brooksong\}@zju.edu.cn}
}

\maketitle              % typeset the title of the contribution

\begin{abstract}
Although deep learning has achieved impressive advances in transient stability assessment of power systems, the insufficient and imbalanced samples still trap the training effect of the data-driven methods. This paper proposes a controllable sample generation framework based on Conditional Tabular Generative Adversarial Network (CTGAN) to generate specified transient stability samples. To fit the complex feature distribution of the transient stability samples, the proposed framework firstly models the samples as tabular data and uses Gaussian mixture models to normalize the tabular data. Then we transform multiple conditions into a single conditional vector to enable multi-conditional generation. Furthermore, this paper introduces three evaluation metrics to verify the quality of generated samples based on the proposed framework. Experimental results on the IEEE 39-bus system show that the proposed framework effectively balances the transient stability samples and significantly improves the performance of transient stability assessment models.
% We would like to encourage you to list your keywords within
% the abstract section using the \keywords{...} command.
\end{abstract}
\section{Introduction}
Power system transient stability assessment is one of the most significant ways to ensure the security and stability of power systems. It assesses the ability of a power system to recover to the original secure state or transition to a new secure state after withstanding a specific disturbance \cite{wu_survey_2012}. Therefore, fast and accurate transient stability assessment is needed to deal with emergencies in time and effectively ensure the secure operation of power systems. However, Time Domain Simulation (TDS), a traditional method of transient stability assessment, is extremely time-consuming due to the nonlinear complexity of power systems \cite{tang_transient_1994}. In recent years, to improve the computational speed of assessment models, several transient stability assessment methods based on deep learning are proposed \cite{gao_transient_2019,li_transient_2020,tacchi_model_2020,wei_real-time_2019}. These assessment methods are usually data-driven, and need large-scale valid samples \cite{bo_power_2016,vasant2018intelligent,vasant2019intelligent,vasant2020intelligent}.

However, there are two problems that need to be addressed when training the assessment model in practice. Firstly, the insufficient samples cannot effectively represent the distribution of features, resulting in the risk of model overfitting. Moreover, since the category distribution of samples is highly imbalanced, the learning of the unstable samples is usually inhibited, leading to poor performance of trained models on unstable samples.

To solve the insufficient and imbalanced samples and improve the performance of transient stability assessment models, we use a sample generation model to supplement the transient stability samples, especially the unstable samples. Generative Adversarial Network (GAN) \cite{goodfellow_generative_2014} is widely used in sample generation tasks, which trains a generator and a discriminator in the adversarial process. However, the generation process of GAN is uncontrollable, resulting in a large number of unnecessary samples. Instead, Conditional GAN (CGAN) realized a conditional generation mechanism based on the architecture of GAN to generate required samples \cite{mirza_conditional_2014}. Furthermore, since transient stability samples are usually recorded as tabular data, we focus on the dedicated CTGAN method which implemented mode-specific normalization and conditional generation for tabular data \cite{xu_modeling_2019}.

Therefore, this paper proposes an imbalanced sample generation framework based on CTGAN for power system transient stability. Considering the structural characteristics of transient stability samples, the generation framework firstly models the samples as tabular data, and uses the Gaussian Mixture Model (GMM) to normalize the tabular data \cite{anzai_2012_pattern,tsukakoshi_analysis_2012}. Multiple conditions, including the transient stability and the load level, are converted into a single condition vector to enable multi-conditional generation. Besides, we design a multi-metric evaluation to effectively evaluate the obtained sample generation framework. The evaluation includes the effect of conditional generation, distance calculation, and the performance of transient stability assessment models trained with generated samples. Case studies on the IEEE 39-bus system show that the proposed framework can effectively balance the transient stability samples and significantly improve the performance of transient stability assessment models. 
\section{Sample Generation Framework for Transient Stability}
In this section, we detail the proposed sample generation framework based on CTGAN for power system transient stability. As shown in Fig. 1, the proposed sample generation framework first models transient stability samples as tabular data, then transform the data using one-hot code and GMM normalization, and finally train the CTGAN model.
\subsection{Transient Stability Sample Representation}
To construct appropriate input characteristics, we should not only consider the correlation between characteristics and transient stability, but also consider whether the characteristics can be obtained in real time or quickly calculated in actual power system. Assuming that there is no another fault in the transient process, the transient stability of the power system has been determined at the moment of fault removal. Therefore, we take the values at the moment of fault clearing as the representation of transient stability samples. A transient stability sample is represented by the voltage magnitude and voltage angle of bus nodes, active power and reactive power of load nodes, active power and reactive power of generator nodes at the moment of fault clearing. 
% figure 1
\begin{figure}
\vspace{-0.6em}
\centering
\includegraphics[scale=0.6]{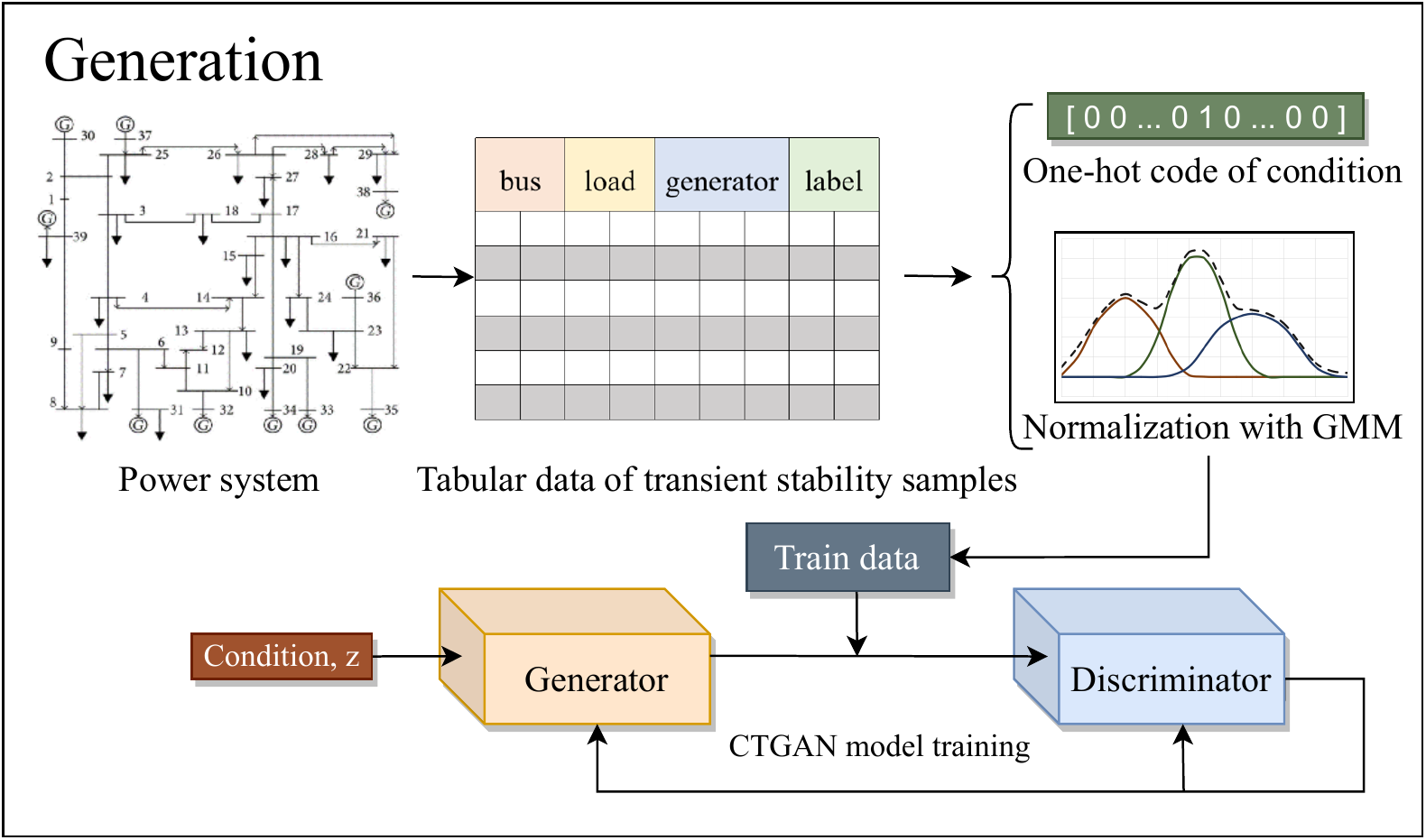}
\caption{Illustration of the proposed imbalanced sample generation framework based on CTGAN for power system transient stability.}
\vspace{-2em}
\end{figure}

% 2.2
\subsection{Transformation of Multi-condition Vector}
With the basic idea of conditional generation, the transient stability and the load level of transient stability samples are used as generation conditions to realize multi-conditional generation. However, in common CGANs, the conditional vector is a one-hot code, which can only represent a single condition. Therefore, a simple transformation method for multi-condition vector is designed in the proposed generation framework, which aims to convert multiple condition vectors into a single condition vector:
\begin{equation}
    cond_1 \oplus cond_2 \oplus \cdots \oplus cond_n
\end{equation}
where $cond$ represents conditions, $n$ represents the number of conditions, and $\oplus$ represents the operation of serially concatenate. The specific principle is that $n$ condition vectors can be serially concatenated, and then transformed into one condition vector as the condition input of CTGAN model.
% \begin{equation}
%     cond_1 \otimes cond_2 \otimes \cdots \otimes cond_n
% \end{equation}
% where $cond$ represents conditions, $n$ represents the number of conditions, $\oplus$ and $\otimes$ represent the operation of serially concatenate and multiply, respectively. The specific principle is that $n$ condition vectors can be serially concatenated or multiplied, and then transformed into one condition vector as the condition input of CTGAN model. In this paper, we serially concatenate the conditions to construct the condition input.

% 2.3
\subsection{Normalization with GMM}
To eliminate the dimensional influence between different characteristics, it is important to transform the samples through appropriate methods before inputting them into the model for training. Transient stability samples are composed of the feature values of bus, load, and generator nodes. However, these continuous values cannot be normalized by one-hot code. 

Considering the complex distribution of transient stability samples, the general min-max normalization is unable to fit the complex distribution. Therefore, when processing transient stability samples, the variational GMM is used to process continuous values to fit the complex distribution of each feature. The basic steps of the normalization are elaborated as follows: 

\subsubsection{Learning GMM.}
For each continuous column $C_i$, we use a variational Gaussian mixture model to learn a GMM distribution: 
\begin{equation}
    \mathbb{P}_{C_i}(c_{i,j}) = \sum_{k=1}^{m_i} \mu_k \mathcal{N} (c_{i,j};\eta_k, \phi_k)
\end{equation}
where $m_i$ is the number of modes, $\mu_k$, $\eta_k$ and $\phi_k$ are weight, mean value and standard deviation of the $k^{th}$ mode, respectively.
\subsubsection{Calculating probability density.}
For each value $c_{i,j}$ in column $C_i$, we calculate the probability density of each mode:
\begin{equation}
    \rho_k = \mu_k \mathcal{N} (c_{i,j};\eta_k, \phi_k)
\end{equation}
\subsubsection{Normalization.}
We find the highest $\rho_k$ in $m_i$ modes and normalizing it. For instance, if the highest probability density $\eta_2$ in three modes $\eta_1, \eta_2, \eta_3$, the value $c_{i,j}$ can be transformed to a one-hot code $[0,1,0]$ and a scalar $\beta_{i,j} = (c_{i,j} - \eta_2)/4\phi_2$ normalized to $[-1,1]$.

% 2.4
\subsection{CTGAN-based Network}
We adopt CTGAN model as the basic sample generation model, which includes a generator and a discriminator. And we construct the generator and the discriminator with fully connected layers respectively. 

The processed transient stability samples are applied as the training input of the constructed CTGAN-based network. In the training process, the discriminator and generator are trained by turns to obtain the model for the sample generation framework. To test the model, we apply it to the generation task of transient stability samples with labels., And we can also control the generating conditions to generate samples with specific labels purposefully, such as controlling the model to generate transient unstable samples.

% 3
\section{Multi-metric Evaluation}
After realizing the generation framework of power system transient stability samples, it is necessary to evaluate the generation framework. This paper designs a multi-metric evaluation for the transient stability samples generation framework. As shown in Fig. 2, the evaluation is composed of the following three metrics: the effect of conditional generation, the distance between real samples and generated synthetic samples, and the performance of assessment models trained with generated samples. 
% figure 2
\begin{figure}
\vspace{-0.6em}
\centering
\includegraphics[scale=0.6]{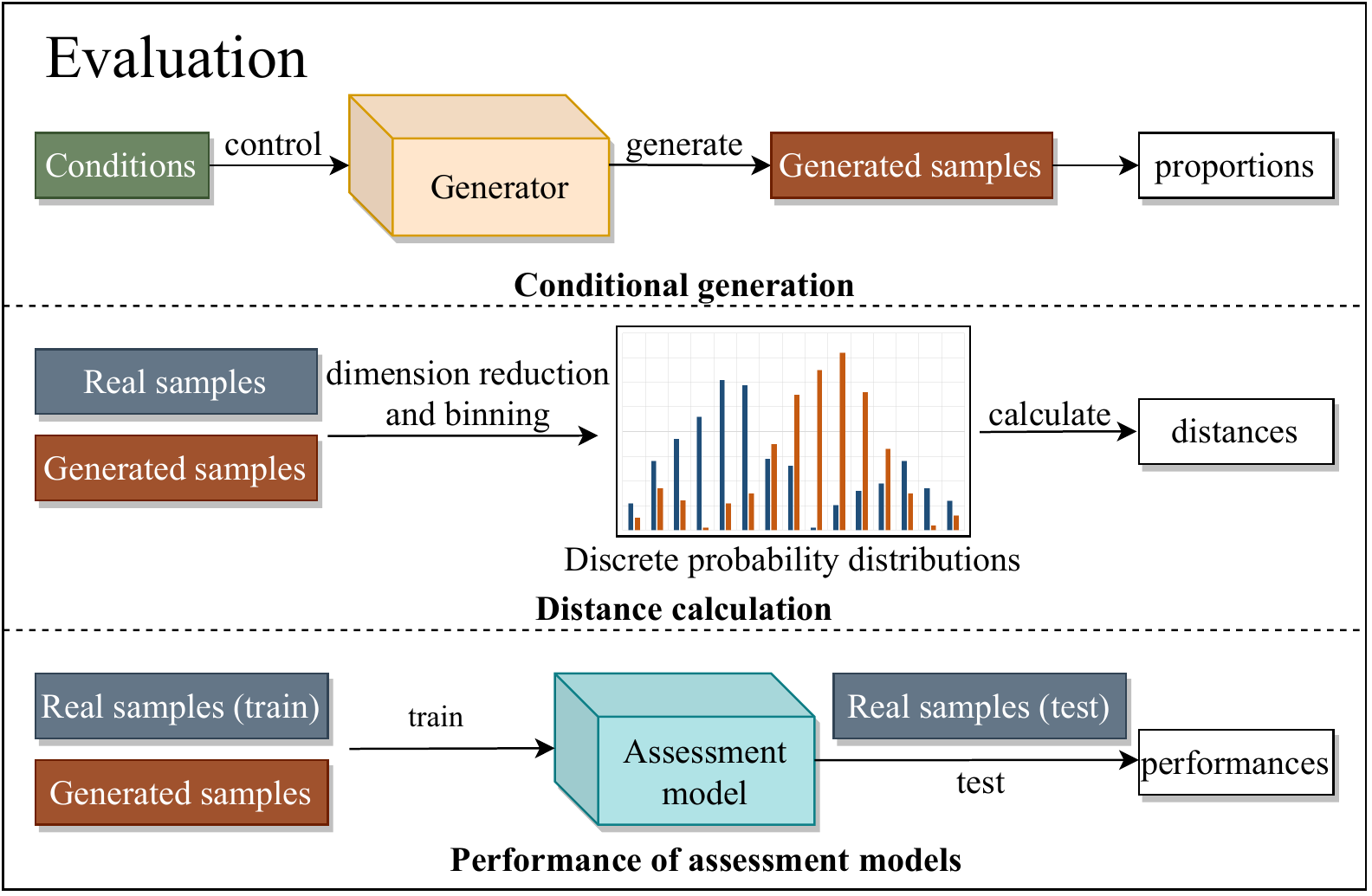}
\caption{Illustration of three evaluation metrics for the sample generation framework based on CTGAN.}
\vspace{-2em}
\end{figure}

% 3.1
\subsection{Conditional Generation}
The power system transient stability sample generation framework should have the ability to control the transient stability and the load level characteristics of power system samples in generating process. By comparing the proportions of transient stability samples that generated under different settings (without setting conditions, setting conditions as transient stable, and setting conditions as transient unstable), the condition generation ability of the transient stability can be evaluated. The same is true for the evaluation of the load level condition.
% 3.2
\subsection{Distance Calculation}
Without setting the generating conditions, the generated samples should be similar to the real samples as much as possible. Therefore, calculating the similarity or distance between the two distributions is an efficient metric for evaluating the generation framework. 

First, dimensionality reduction methods, such as Principal Component Analysis (PCA) \cite{yata_principal_2015}, should be used to reduce the dimension of the samples to some appropriate degree. Second, we convert the dimensionality reduced samples into discrete probability distributions through the binning operation. Finally, we calculate the distance between the probability distribution of synthetic samples and real samples. Common methods for measuring the similarity between two distributions are adopted to calculate the distance between the distributions, such as KL divergence, JS divergence, and Wasserstein distance \cite{arjovsky_wasserstein_2017}.
% \cite{kullback_information_1951,lin_lin_1991,arjovsky_wasserstein_2017}.
% 3.3
\subsection{Performance of Assessment Models}
To evaluate the generated samples more practically, the performance of the transient stability assessment model trained with generated samples is a proper metric. Some classical networks for classification are selected as the power system transient stability assessment model, the generated samples are used for the training of assessment models, and the performance is obtained by testing the assessment models. 

More specifically, the real dataset of transient stability samples is randomly divided into $S_{train}$ and $S_{test}$. We randomly generate $S_{gen}$ and get the united set $S_{union} = S_{train} + S_{gen}$. And $S_{train}$, $S_{gen}$, and $S_{union}$ are used for the training of the transient stability assessment models respectively to obtain different assessment models and the models are tested on $S_{test}$. These models are tested on the real test set to obtain the accuracy, recall rate of transient stable samples, recall rate of transient unstable samples. And these test scores can be used as the evaluation metric to evaluate the quality of the generation framework.

% 4
\section{Experiment}
In this section, we study our proposed framework on the classical IEEE 39-bus power system \cite{pai_energy_1989} and show its excellent performance by evaluating the effect of conditional control, calculating the distance between distributions and the scores of assessment models trained with generated samples.
% 4.1
\subsection{Experimental Setup}
\subsubsection{Time Domain Simulation Samples.}
Matpower \cite{zimmerman_matpower-matlab_1997} and Power System Analysis Toolbox (PSAT) \cite{ayasun_voltage_2006} are applied to obtain the original dataset of real samples, taking the IEEE 39-bus system as the basic system. The power system contains 39 buses, 10 generators, 19 loads and 46 transmission lines. For simulating the transient stability samples, we adopt the following principles:
\begin{enumerate}
    \item Randomly changing both active and reactive power of all loads from 60\% to 145\% of basic load level.
    \item Using the matpower to compute the optimal power flow for the next TDS.
    \item Randomly selecting a fault line, setting a three-phase grounding fault from 20\% to 80\% and clearing it after a time from 1/60 to 1/3 seconds.
    \item Using the PSAT to do time domain simulation for 10 seconds.
    \item Labeling the stability of generated sample by values of generators after TDS.
\end{enumerate}
With the simulation operations above performing, we generate a total of 14,221 transient stability samples that include 11,510 stable samples and 2,711 unstable samples as the original dataset.
\subsubsection{Generation Model Training.}
CTGAN is used as the primary sample generation model, which includes a generator and a discriminator.. In the generator, two fully connected layers are used, and each fully connected layer is equipped with a batch normalization layer and a ReLU activation layer. The tanh and softmax activation functions are used for the output layer. In the discriminator, two fully connected layers are used, and the dropout layer is used to filter the nodes appropriately to reduce overfitting. 

% 4.2
\subsection{Evaluation Metrics}
The CTGAN-based generation framework of power system transient stability samples is trained with the simulated samples as the training set. After that, it is necessary to evaluate the quality of the generation framework. This paper designs a multi-metric evaluation for the generation framework of transient stability samples, composed of three evaluation metrics. 
\subsubsection{The Effect of Conditional Generation.}
We evaluate the ability to control the transient stability and the load level of power system samples in generating. 

Table 1 shows the result of conditional generation with different transient stability condition settings. We set the conditions as follows: no condition, stable, and unstable. When the condition is set as transient stable, the proportion of stable samples generated is increased by 18.7\% compared with the samples generated without condition. When the condition is set as unstable, the increment is 48.8\%. The result shows that the transient stability ratio of the generated samples can be effectively controlled, and the framework can effectively balance the transient stability samples by generating more unstable samples.
% table 1
\begin{table}
\caption{The result of generation with different transient stability conditions.}
\begin{center}
\setlength{\tabcolsep}{3mm} % 表格过窄
\begin{tabular}{ccc}
% \hline
% \multicolumn{1}{l}{\rule{0pt}{12pt}Condition} 
% & \multicolumn{2}{l}{Stable proportion (\%)}
% & \multicolumn{3}{l}{Unstable proportion (\%)}\\[2pt]
\hline\rule{0pt}{12pt}
Condition &	Stable proportion (\%) &	Unstable proportion (\%) \\[2pt]
\hline\rule{0pt}{12pt}
Without condition         &	59.92   & 40.08 \\
With condition (stable)   &	71.10   & 28.90 \\
With condition (unstable) &	40.38   & 59.62 \\[2pt]
\hline
\end{tabular}
\end{center}
\vspace{-2em}
\end{table}

Moreover, the result of conditional generation with different load level condition settings is shown in Table 2. We set the conditions as no condition, and as 18 load levels (60\% to 145\%, with a step of 5\%). We count the number of samples of corresponding load level in the generated samples under the control of generation conditions, and calculate the proportion for comparison. When the condition is set to a specific load level, the proportion of the corresponding load level generated will be higher than that of the samples generated without condition. The results show that the generation framework can effectively control the load level proportion of the generated samples.

% table 2
\begin{table}
\caption{The result of generation with different load level conditions.}
\begin{center}
\setlength{\tabcolsep}{3mm} % 表格过窄
\begin{tabular}{cccc}
\hline\rule{0pt}{12pt}
\makecell[c]{Condition} &	\makecell[c]{Generated without\\ condition (\%)} &	\makecell[c]{Generated with load\\ level condition (\%)} &	\makecell[c]{Rate of\\ improvement (\%)} \\[2pt]
\hline\rule{0pt}{12pt}
70\%  &	2.49 &	3.35 &	34.54 \\     
80\%  &	3.89 &	4.64 &	19.32 \\      
90\%  &	2.70 &	4.92 &	81.90 \\  
100\% &	3.38 &	4.59 &	36.09 \\ 
110\% &	5.81 &	8.86 &	52.60 \\ 
120\% &	2.51 &	2.58 &	2.67  \\ 
130\% &	2.50 &	4.60 &	84.53 \\ 
140\% &	0.54 &	1.81 &	233.76\\ [2pt]
\hline
\end{tabular}
\end{center}
\vspace{-2em}
\end{table}

\subsubsection{The Distance between Real and Generated Sample Distribution.}
The generated samples should be similar to the real samples as much as possible. Therefore, calculating the distance between the two distributions is an efficient metric for evaluating the generation framework. 

Table 3 shows the results of JS divergence and Wasserstein distance calculated between distributions. We randomly select 2,000 samples from real samples as set $S_{real}^A$, repeat the operation to get $S_{real}^B$, and generate 2,000 samples as set $S_{gen}$. From Table 3, we can see that the distance between $S_{real}^A$ and $S_{gen}$ and the distance between $S_{real}^A$ and $S_{real}^B$ are in the same order of magnitude, which means that the samples generated by the generation framework are similar to the real samples in these three distance measurements.
% table 3
\begin{table}
\caption{The distance between distributions.}
\begin{center}
\setlength{\tabcolsep}{3mm} % 表格过窄
\begin{tabular}{ccc}
\hline\rule{0pt}{12pt}
\makecell[c]{Distributions} & \makecell[c]{JS divergence} & \makecell[c]{Wasserstein distance} \\[2pt]
\hline\rule{0pt}{12pt}
$S_{real}^A$, $S_{real}^B$ & 0.002826 &	0.001429 \\[2pt]
$S_{real}^A$, $S_{real}^B$ & 0.063141 & 0.006388 \\[2pt]
$S_{real}^A$, $S_{real}^B$ & 0.063084 & 0.005939 \\[2pt]
\hline
\end{tabular}
\end{center}
\vspace{-2em}
\end{table}

\subsubsection{The Performance of Assessment Models Trained with Generated Samples.}
The performance of assessment models trained with generated samples is a valuable metric. In this paper, we select Multilayer Perceptron (MLP) and Decision Tree (DT)
% Multilayer Perceptron (MLP) \cite{mw_artificial_1998} and Decision Tree (DT) \cite{safavian_survey_1991} 
as the power system transient stability assessment models for training and testing, since they are classical network models for classification problems. The hidden layer size of MLP is 200, and the max number of iterations is 500. The max depth of DT is 100.

Table 4 shows the test results of assessment models trained with different datasets. We randomly divide the real dataset into $S_{train}$ with 8,533 samples and $S_{test}$ with 5,688 samples. We randomly generate $S_{gen}$ with 8,533 samples and get the united set $S_{union} = S_{train}+S_{gen}$. Note that $S_{train}$, $S_{gen}$, and $S_{union}$ are used for the training of the transient stability assessment models respectively to obtain different assessment models and the models are tested on $S_{test}$. The scores of the model trained with $S_{gen}$ are lower than the scores of the model trained with $S_{train}$. However, the scores of the model trained with $S_{union}$ are higher than the scores of the model trained with $S_{train}$. The recall rate of unstable samples is increased by 1.48\% in DT, and is increased by 2.74\% in MLP. The results show that adding the generated samples into the train set is able to improve the performance of transient stability assessment models, especially for the unstable label, which is the scarce class in the train set.
% table 4
\begin{table}
\caption{The test results of assessment models trained with different datasets. Recall$_P$ is the recall rate of stable samples, and Recall$_N$ is the recall rate of unstable samples.}
\begin{center}
\setlength{\tabcolsep}{3mm} % 表格过窄
\begin{tabular}{clcccc}
\hline\rule{0pt}{12pt}
\makecell[c]{Assessment \\ model} &	\makecell[c]{Train \\ dataset} &	\makecell[c]{Recall$_P$} &	\makecell[c]{Recall$_N$} & \makecell[c]{F1 score} & \makecell[c]{Accuracy} \\[2pt]
\hline\rule{0pt}{12pt}
DT  & $S_{train}$  & 0.9770 & 0.9348 & 0.9813 & 0.9694 \\
DT  & $S_{gen}$    & 0.9430 & 0.6856 & 0.9376 & 0.8969 \\
DT  & $S_{union}$  & 0.9788 & 0.9486 & 0.9837 & 0.9734 \\
MLP & $S_{train}$  & 0.9883 & 0.9261 & 0.9861 & 0.9771 \\
MLP & $S_{gen}$    & 0.7719 & 0.8061 & 0.8509 & 0.7780 \\
MLP & $S_{union}$  & 0.9832 & 0.9515 & 0.9863 & 0.9775 \\ [2pt]
\hline
\end{tabular}
\end{center}
\vspace{-2em}
\end{table}

% 5
\section{Conclusion}
In this paper, we attempt to solve the imbalanced distribution and insufficient samples in the research of power system transient stability assessment. We propose a CTGAN-based controllable sample generation framework for transient stability. In the generation framework, firstly, the transient stability samples are processed into tabular data. Then the transient stability and load level are converted into the conditional vector and the variational Gaussian mixture model is used to fit and normalize the tabular data. And finally train the CTGAN model with processed samples. Moreover, we design a multi-metric evaluation to effectively evaluate the generation framework from three aspects: the effect of conditional generation, the distance between real and generated sample distribution, and the performance of the assessment model trained with generated samples. Experiments demonstrate that samples generated through the proposed generation framework are valid and effective in multiple metrics. 

\subsubsection{Acknowledgement.}
This work is funded by National Key Research and Development Project (Grant No: 2018AAA0101503) and State Grid Corporation of China Scientific and Technology Project: Fundamental Theory of Human-in-the-loop Hybrid-Augmented Intelligence for Power Grid Dispatch and Control.


\bibliographystyle{splncs03_unsrt}
\bibliography{ref1}

\end{document}